\documentclass[10pt,twocolumn,letterpaper]{article}

\usepackage{cvpr}
\usepackage{times}
\usepackage{epsfig}
\usepackage{graphicx}
\usepackage{amsmath}
\usepackage{amssymb}
\usepackage{graphicx}
\usepackage{subcaption}
\usepackage{xcolor}
\graphicspath{ {.} }

\cvprfinalcopy %
\def\cvprPaperID{****} 

\begin{document}

\title{
Visual Localization Using Semantic Segmentation and Depth Prediction
}

\author{
Huanhuan Fan\thanks{Equal contribution.} \quad
Yuhao Zhou\footnotemark[1] \quad
Ang Li\thanks{Corresponding author.} \quad
Shuang Gao \quad
Jijunnan Li \quad
Yandong Guo \\
OPPO Research Institute\\
\tt\small \{zhouyuhao, fanhuanhuan, liang, gaoshuang, lijijunnan, guoyandong\}@oppo.com
}

\maketitle

\begin{abstract}

In this paper, we propose a monocular visual localization pipeline leveraging semantic and depth cues.
We apply semantic consistency evaluation to rank the image retrieval results and a practical clustering technique to reject estimation outliers.
In addition, we demonstrate a substantial performance boost achieved with a combination of multiple feature extractors.
Furthermore, by using depth prediction with a deep neural network, we show that a significant amount of falsely matched keypoints are identified and eliminated.
The proposed pipeline outperforms most of the existing approaches at the Long-Term Visual Localization benchmark 2020 \footnote{https://www.visuallocalization.net/}. 
\end{abstract}

\section{Introduction}


Visual localization is one of the major challenges and the fundamental capability for numerous computer vision applications, such as augmented reality, intelligent robotics, or long-term visual navigation of autonomous driving\cite{lim2012real}\cite{Castle2008Video}.
A core task for these applications is to find the precise location of any query image with a map of 3-dimensional (3D) points reconstructed by structure-from-motion (SfM) with database images\cite{wang2015adaptive}\cite{schonberger2016structure}\cite{ullman1979interpretation}. 
The map is typically used to describe the position of landmarks\cite{fuentes2015visual}\cite{valgren2010sift}, i.e., 3D points and structures in the environments, which are pre-collected from the point features extracted from the database images.
Every point feature is associated with a 3D position in the world coordinates and a unique descriptor of its visual appearance.
During localization, 
the points in the 2D query image and the points with 3D world coordinates
in the map is determined, and finally the 6-Degree-of-Freedom (DoF) pose of query image is recovered with Perspective-n-Point (PnP)\cite{gao2003complete}.

There are several critical challenges associated with the repeatability of traditional feature descriptors including SIFT\cite{ng2003sift} and BRIEF\cite{valgren2010sift}, which are shown to be highly sensitive to large variation in viewpoint and illumination\cite{revaud2019r2d2}\cite{detone2018superpoint}\cite{yi2016lift}\cite{dusmanu2019d2}. Consequently, localization pipelines struggle to find enough 2D-3D matched pairs to facilitate the successful 6DoF pose estimation.
In addition, our experimental results show that keypoints are often incorrectly matched even after a standard RANSAC-based outlier rejection mechanism\cite{mach1981random}. 

In this paper, we propose the following key enhancement on the retrieval-based visual localization pipeline to handle the problems.
\textbf{(i)}
We evaluate the confidence of every image retrieval result with the proposed \textit{semantic consistency weight} (SCW), which ranks the retrieved results based on the ratio of matched keypoints with consistent semantic label. 
\textbf{(ii)}
The outliers of pose estimates are identified by a practical clustering technique. 
\textbf{(iii)}
We demonstrate the effectiveness of jointly using multiple learning-based features and descriptors for improved 6DoF pose estimation.
\textbf{(iv)}
The enhanced R2D2 descriptor enables higher matching accuracy. 
\textbf{(v)}
False keypoint matching relations are identified with the proposed \textit{semantic consistency check} (SCC) and \textit{depth consistency verification} (DCV) procedure.
\textbf{(vi)}
We leverage semantic segmentation with the proposed weighted-RANSAC scheme that adaptively adjusts the RANSAC threshold.

\section{Method}

\subsection{Overview}

\begin{figure*} [t]
\includegraphics[width=18cm]{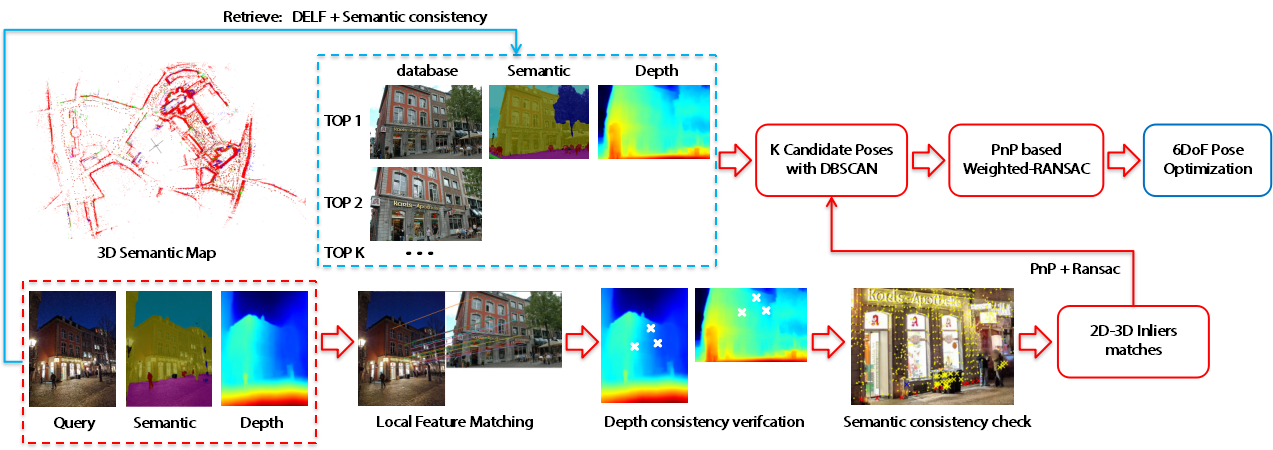}
\centering
\caption{
An overview diagram of the proposed pipeline. 
}
\label{fig:overview}
\end{figure*}

Long-term visual localization is a challenging task due to weak textures and potentially huge variation in viewpoint and illumination.
Figure \ref{fig:overview} illustrates the proposed localization pipeline based on a standard retrieval-based framework in\cite{irschara2009structure}, 
which consists of three main modules as follows.

\textbf{(i) Map construction.}
With database images captured at the target scene, we first run a standard SfM algorithm to construct a 3D model, represented by a large number of 3D interest points. 
At the same time, the 2D-3D keypoint matching relations produced during SfM are recorded for run-time pose estimation.
Moreover, semantic segmentation is estimated for every database image and semantic labels are associated with the 3d points.

\textbf{(ii) Image retrieval.}
A coarse search is performed by matching the query with the database images using global descriptors computed by a deep neural network reported in \cite{noh2017large}. 
Though this process is efficient given that there are far fewer database images than the 3D points in SfM model, it often produces incorrect results, especially under large variation of viewpoint and illumination.
We handle this problem by first calculating homography matrices with existing matched pairs, and cluster the results to reject outliers that correspond to incorrect retrieval results. 
Furthermore, we propose SCW, which reranks the retrieval results with a verification step considering semantic consistency.
As a result, a number of candidate images are discovered that are semantically consistent to the query image. 

\textbf{(iii) Feature matching and pose estimation.}
2D interest points are extracted from query and matched against keypoints in every candidate image, 2D-3D query-database matching relations are built for PnP pose estimation. 
Nevertheless, our experimental results show that incorrect matches are often produced despite of the RANSAC mechanism. 
This issue is handled by the following sub-modules.
Firstly, a combination of multiple learning-based feature extractors is used which produces substantially more feature points.
Secondly, SCC and DCV are performed for semantic and depth consistency verification, which essentially remove weakly consistent matches.
Finally, all matching relations are weighted with consistency scores for bias sampling during the novel weighted-RANSAC and PnP.

\subsection{Semantic verification of image retrieval}
\label{sec:semantic}

\begin{figure}[t]
\centering
\includegraphics[width=8.5cm]{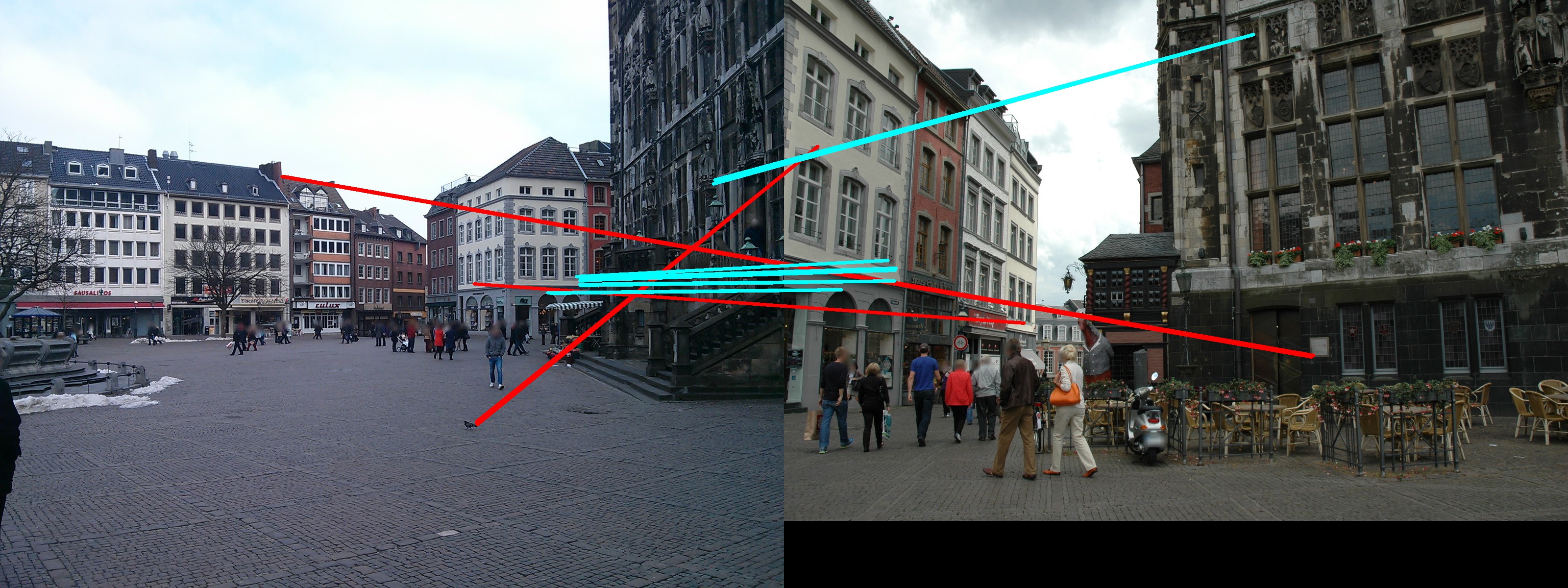}
\caption{
Experiment results on SCC: matching results of KNN+RANSAC in red and KNN+SCC+RANSAC in cyan.
}
\label{fig:scc_1}
\end{figure}

Recently, learning-based approaches achieve impressive performance on image retrieval\cite{gordo2016deep}\cite{teichmann2019detect}\cite{noh2017large}\cite{arandjelovic2016netvlad}\cite{babenko2015aggregating}. Our experimental results show that DEep Local Feature (DELF)\cite{noh2017large} produces the best results.
Therefore, we apply DELF to obtain a predefined number of candidate images.
Nevertheless, a significant amount incorrect results are inevitably generated.
This problem is handled by the proposed SCW and a clustering process.
Current state-of-the-art semantic segmentation approaches are mainly based on deep neural networks, such as PSPNET\cite{zhao2017pyramid}, BiSeNet\cite{yu2018bisenet} and DeepLabv3+\cite{chen2018encoder}, where DeepLabv3+ is selected for out pipeline that attains superior performance on PASCAL VOC2012\cite{everingham2010pascal} and Cityscapes\cite{cordts2016cityscapes}.
For every retrieved image, the homography matrix from the query image is computed by RANSAC, producing a number of inliers, denoted by $S_{c}$. We also count the number of matches with identical semantic label represented by $S_{f}$.
SCW is then defined by
\begin{equation} 
\label{eq:rerank}
S_{R} = \cfrac {\alpha_{1} S_{c} + \alpha_{2} S_{f}} {\alpha_{1} + \alpha_{2}} ,
\end{equation}
where $\alpha_{1}$ and $\alpha_{2}$ are scalars associated with scoring weights. SCW is then calculated for all query-retrieval pairs, and only those with high SCW scores will proceed to subsequent steps.
Due to inaccurate semantic prediction, a number of incorrect candidate images may still present. This issue is tackled with a standard clustering technique that effectively rules out the outliers of pose estimates produced by the problematic retrieval results.

Semantic consistency is also computed to eliminate the inconsistent matches with the proposed SCC, described as follows. 
For an output matched pair of features, the semantic labels of both query point and retrieved point are compared, and pairs with different labels are considered as unsuccessful and thus discarded.
In addition, features points on dynamic objects including pedestrians and vehicles are also removed. 
Figure \ref{fig:scc_1} shows a comparison of the matching results before and after introducing SCC, where incorrect matches are eliminated. Note that SCC is applied before RANSAC. 

\subsection{Multiple feature extractors and enhanced R2D2 descriptor}

Recently, learning based feature extractors and descriptors are shown to outperform traditional handcrafted ones including SIFT \cite{trzcinski2012learning} \cite{mishchuk2017working} \cite{ono2018lf}. 
Learning-based feature extractors are trained on large datasets and are able to automatically discover feature extraction process and representation most suited to the data. SuperPoint\cite{detone2018superpoint} and R2D2\cite{revaud2019r2d2} are amongst the most recent approaches that achieve impressive performance. 
While SuperPoint\cite{detone2018superpoint} is trained to extract multi-scale SIFT-like salient features, R2D2\cite{revaud2019r2d2} focuses on extracting discriminant features with low repeatability, i.e. avoid repeating features such as windows. 
Our experimental results show that a simple combination of features by both detectors results in a significant increase in the number 
of matched pairs, which in turn leads to higher pose estimation accuracy.

Furthermore, an in-depth study on the R2D2 algorithm shows that the confidence of a matching result is closely related to the similarity between the descriptor score of the two features. In other words, a pair of matched keypoints with similar descriptor score is more likely to be true than ones with very different score.
Thus, the proposed modification on R2D2 descriptor involves incorporating the detection confidence score by
\begin{equation} \label{eq:r2d2}
d^\prime = d \times s,
\end{equation}
where $d$ is the original descriptor, $s$ is the descriptor score and $d^\prime$ is the new descriptor with proposed modification.

\begin{figure} [t]
\includegraphics[width=8.5cm]{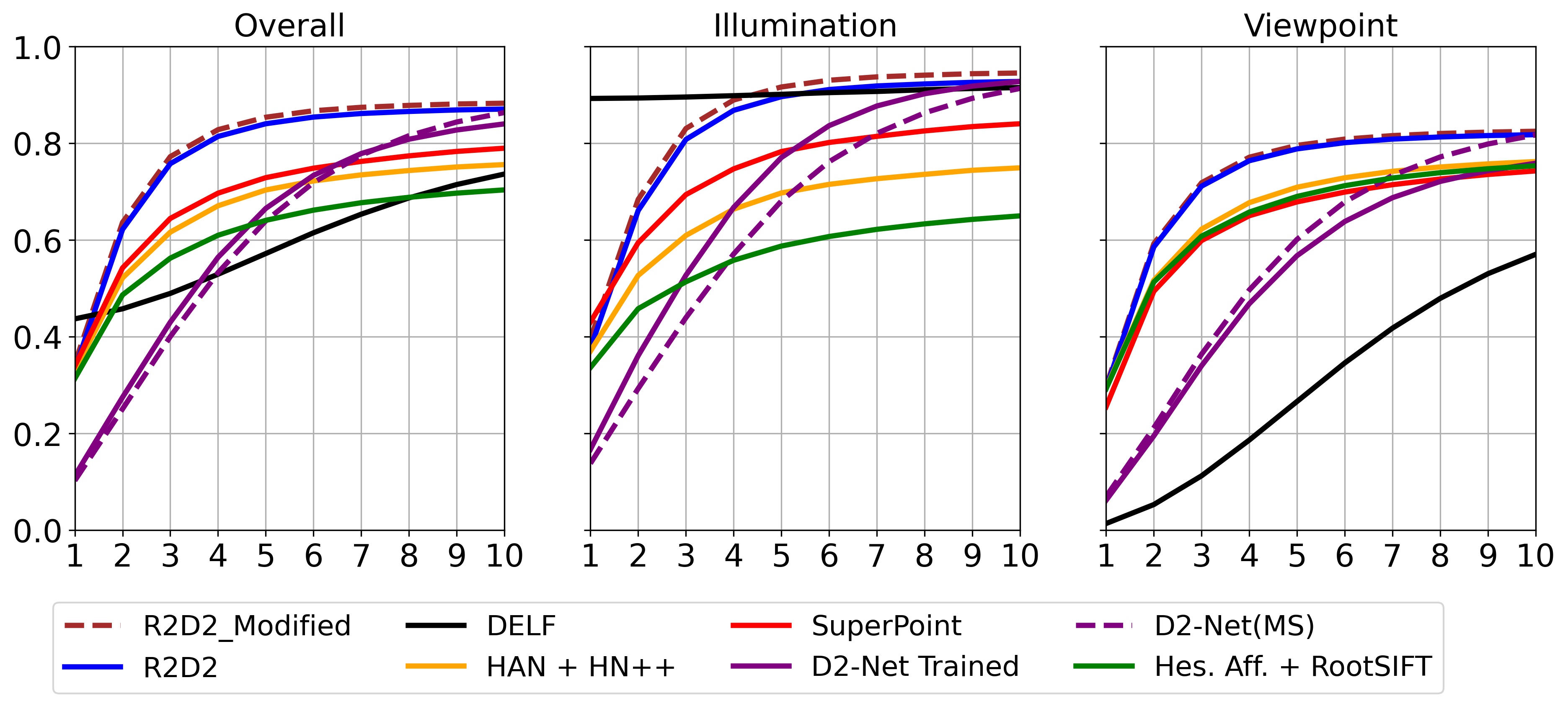}
\centering
\caption{
Comparion with the original R2D2 descriptor and other state of the art using the MMA (vertical axis) for varying reprojection error thresholds (horizontal axis, in pixel) on the HPatches dataset. 
}
\label{fig:r2d2_mod}
\end{figure}

In order to allow realistic and reliable comparisons in different environments, we tested the modified R2D2 descriptor on the HPatches dataset\cite{balntas2017hpatches}, the one of the most recent ones for evaluation of different feature descriptors.
The performance of the proposed modification is evaluated with the mean matching accuracy (MMA), and compared with other latest descriptors including the original R2D2, DELF\cite{noh2017large}, HardNet++ descriptors with HesAffNet regions\cite{mishchuk2017working}\cite{mishkin2018repeatability} (HAN + HN++), SuperPoint\cite{detone2018superpoint}, D2-Net\cite{fathy2018hierarchical} and a handcrafted Hessian
affine detector with RootSIFT descriptor\cite{perd2009efficient}.
Figure \ref{fig:r2d2_mod} shows the results for illumination and viewpoint changes, as well as the overall performance on the HPatches dataset, which indicate a substantial improvement with our modification.

\subsection{Depth consistency verification}
\label{sec:dcv}

In addition to semantic segmentation, recent advances in monocular depth prediction also enables a strong depth constraint on feature matching for accurate visual localization. In this section, we propose DCV which effectively identify fallacious matched features associated with low depth consistency.
SfM produces a cloud of 3D points with a set of database images, each one of which corresponds to a specific feature point in database image \cite{schonberger2016structure}. After feature matching is completed between 2D points in query image and the correctly retrieved database image, the correspondence between the query feature points and 3D points is built. Assuming this is found correctly, the depth estimate of one of the feature points is found by projecting the corresponding 3D point to image plane:
\begin{equation} \label{eq:depth_1}
P_{img} = K (R P_{wrd} + T),
\end{equation}
where $P_{wrd}$ is the coordinate of 3D point, $P_{img}$ is the image coordinates, $K$ is the intrinsic matrix, $R$ is the estimated rotation matrix and $T$ is the estimated translation matrix with PnP. With known $x$ and $y$ coordinates of the feature point, the scale of $P_{img}$ can be determined hence its $z$ coordinate, or the \textit{estimated depth value} (EDV). 
This allows us to check the depth value against the one obtained from a deep network such as the work reported in \cite{li2018megadepth}, or the \textit{predicted depth value} (PDV).

For ease of explanation, let us first consider an \textit{ordinary ordinal cost}. Since the scale of the PDV is often unknown, it cannot be compared directly with EDV. One possible solution is to compare their ordinal value.
However, the depth values of different keypoints are not evenly spaced in most cases, and this leads to out-of-scale.
To overcome this issue, we calculated the \textit{adaptive ordinal cost} (AOC) by:
\begin{equation} \label{eq:depth_2}
C = abs(D[i] - D[j]),
\end{equation}

\noindent
where $C$ is the AOC, $D$ is the PDV, and $i$ and $j$ are the ordinal values in the EDVs and the PDVs respectively. 
Assuming the predicted depth is considerably accurate, the AOC scales well for different value intervals. After AOC is estimated for all feature points, they are divided by their mean to cope with the unknown scale.

\begin{figure} [t]
\includegraphics[width=8.5cm]{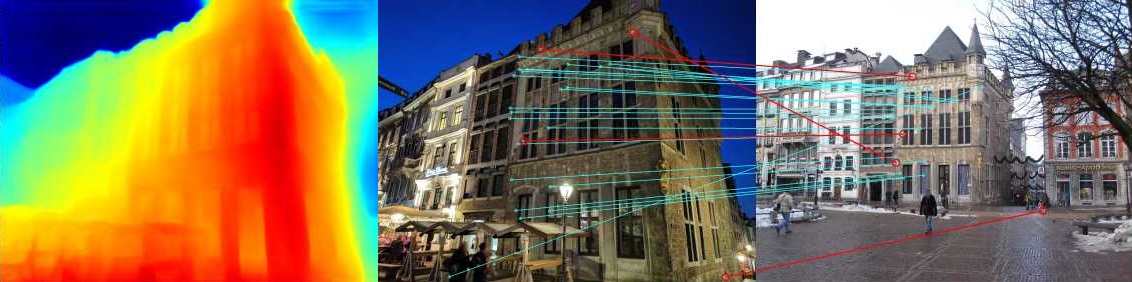} \quad
\centering
\caption{Experiment results showing the identified inconsistent matches: depth prediction in jet-map is on left and matched features on the right, where consistent and inconsistent matches are shown in cyan and red respectively.}
\label{fig:depth_2}
\end{figure}

The effectiveness of the proposed DCV is shown in Figure \ref{fig:depth_2}, where inconsistent matches are clearly identified.
Though there are some false detection due to the inevitable inaccuracy in 3D map generation and monocular depth prediction, they are further neglected by comparing the reprojection error before and after being removed. Nevertheless, a significant portion of inconsistent matches are identified which seriously affect the estimation accuracy.

\subsection{Pose estimation with weighted-RANSAC}

After incorrect matches are removed by SCC and DCV in per query-retrieval pair, 
an overall set of matches are taken between 2D keypoints in the query image and 3D points corresponding to all the retrieved images, allowing the final PnP pose computation. A standard RANSAC PnP compute a hypothetical pose $M$ with a subset of input matches, and all the other samples are tested by computing reprojection error $e_p$ with the pose. A sample $p$ is accepted as an inlier if the error is bellow a preset threshold $e_t$, shown by the following indicator function:
\begin{equation} \label{eq:ransac}
T (p, M) = 
\begin{cases}
1, \quad e_p < e_t \\
0, \quad otherwise.
\end{cases}
\end{equation}
This process is repeated for a number of random samples until enough inliers are found. 
We again leverage semantic information by the proposed weighted-RANSAC scheme, where the threshold is reduced according to semantic consistency, defined by the new indicator function:
\begin{equation} \label{eq:w-ransac}
weight_{\mu} T (p, M) = 
\begin{cases}
1 - (\cfrac {e_p} {\mu \cdot e_t})^{2}, \quad \mid e_p \mid < (\mu \cdot e_t) \\
0, \quad otherwise,
\end{cases}
\end{equation}
where $\mu$ is the reduction ratio calculated by normalizing SCW that is defined in Eqn. \ref{eq:rerank}.

\section{Conclusion}

In this paper, we proposed a visual localization pipeline leveraging semantic and depth cues.
We propose SCW to remove erroneous image retrieval results, SCC to reduce the number of incorrect image retrieval results based on semantic consistency, and DCV to remove matches that have poor depth consistency.
We use the proposed enhanced-R2D2 descriptor and the combination of R2D2 and SuperPoint feature extractor to demonstrate their effectiveness for higher localization accuracy and robustness.
Finally, the outliers of pose estimates are removed with clustering and the final pose estimation is performed by a weighted-RANSAC scheme.
Extensive experimental results show that our method achieves the state-of-art performance on the benchmark.

\bibliographystyle{unsrt}
\bibliography{vlsbib}

\end{document}